\documentclass[accepted]{ci4ts2024} 
                        
\usepackage[american]{babel}

\usepackage{natbib} 
\usepackage{mathtools} 
\usepackage{booktabs} 
\usepackage{tikz} 

\usetikzlibrary{automata,positioning,fit}

\title{Industrial-Grade Time-Dependent Counterfactual Root Cause Analysis through the Unanticipated Point of Incipient Failure: a Proof of Concept}

\author[1]{\href{mailto:<alexandre.trilla@alstomgroup.com>?Subject=UAI 2024 paper}{Alexandre Trilla}{}}
\author[2]{Rajesh Rajendran}
\author[3]{Ossee Yiboe}
\author[4]{Quentin Possama\"{i}}
\author[3]{Nenad Mijatovic}
\author[5]{Jordi Vitri\`a}
\affil[1]{%
    Alstom, Barcelona (Spain)
}
\affil[2]{%
    Alstom, Bengaluru (India)
}
\affil[3]{%
    Alstom, Saint Ouen (France)
}
\affil[4]{%
    Alstom, Villeurbanne (France)
}
\affil[5]{%
    Universitat de Barcelona, Barcelona (Spain)
}

\begin{document}
\maketitle

\begin{abstract}
This paper describes the development of a counterfactual
Root Cause Analysis diagnosis approach for an industrial 
multivariate time series environment. It drives the attention 
toward the Point of Incipient Failure, which is the moment
in time when the anomalous behavior is first observed, and 
where the root cause is assumed to be found before the
issue propagates. The paper presents the elementary
but essential concepts of the solution
and illustrates them experimentally on a simulated 
setting. Finally, it discusses avenues of improvement for the
maturity of the causal technology to meet the robustness
challenges of increasingly complex environments in the industry.
\end{abstract}

\section{Introduction}
\label{secIntro}
The degradation of complex industrial assets is a multifaceted problem that
can be explained by different factors. For instance, in the
Reliability Engineering field, assets are most expected to fail
either prematurely (early) during their break-in period, or late by
the end of their remaining useful life (wear-out)~\citep{Dersin23}.
These failure types can be anticipated because their modes and 
mechanisms are well known. Moreover,
their impact can be mitigated by introducing quality checks in the
manufacturing process and inspection actions in their 
(more-or-less conservative) preventive maintenance schedule.
However, for as long as the machines operate, failures can randomly 
appear at any point in time. This is especially challenging for
dependable assets while they transit the middle region, when the
failure rate is relatively low, but uniform/constant.

In this uncertain setting, the field of Predictive Maintenance
tackles the problem by introducing the data as a means to closely
follow the actual degradation of each asset and make better
informed and timely decisions~\citep{Fink20}.
In this sense, the detection of anomalous
behaviors and the capacity to \emph{diagnose} their 
\emph{root causes} become
increasingly important to guarantee the availability of the
machines. Since these failures appear abruptly, they cannot be
anticipated, and their evolution is not smooth as they
undergo various stages of severe degradation. What is more,
the available operational-service data always comes from the 
field, and thus it is regarded as observational time series 
data, where the value of the variables is always determined by their 
causes, not through experimentation.

The literature on causal approaches for identifying the root 
causes dealing with such type of data typically assume that 
the influences between observed processes change smoothly over 
time, which is introduced as a general confounding 
variable~\citep{Huang15}, and modeled probabilistically to
logically reason on the likelihood of an event within a certain 
interval~\citep{VanHoudt22}. The most common approach, though,
is to frame the problem around the topic of anomaly detection by
modeling the normal operational regime. Specifically,
\cite{Assaad23} develop a summary graph and apply a 
decomposition into abstract causal relations, 
\cite{Budhathoki22} assess the contribution of each variable
to the target outlier score using counterfactuals, 
\cite{Strelnikoff23} develop a
flexible neural graph for edge attribution, 
\cite{Yang22} focus on abnormal data
points that do not follow the regular data-generating process,
and for \cite{Han23}, anomalies are caused by external 
interventions on the normal causal mechanism and therefore
find the algorithmic recourse via counterfactuals
to revert them.

This workshop paper exploits the fact that root causes can be found 
directly from the causal graph and from the time of appearance
of anomalies~\citep{Assaad23}. We follow previous works on
developing a complete causal approach, from the structured model 
building stage to its exploitation through probabilistic 
counterfactual analysis, considering the idiosyncrasies of 
industrial multivariate time series environments, and addressing 
the research gap to learn the system condition \emph{transition}
instead of detecting anomalies.
The paper is organized as follows: Section~\ref{secBackground}
reviews the fundamentals of Root Cause Analysis and
Causal Inference, Section~\ref{secMethod} describes the
proposed causal diagnosis method from the standpoint of 
Predictive Maintenance, Section~\ref{secResults} illustrates
the implementation of the method on a synthetic system,
Section~\ref{secDisc} discusses avenues of improvement
to increase the robustness of the approach, and
Section~\ref{secConc} concludes the work.

\section{Background}
\label{secBackground}

\subsection{Root Cause Analysis}
\label{subRCA}
Root Cause Analysis (RCA) is a troubleshooting method of 
problem solving used for identifying the root causes of faults 
or failures~\citep{Wilson93}.
RCA is a form of deductive inference since it requires an 
understanding of the underlying causal mechanisms for
the potential root causes 
and the problem, i.e., what is typically found in the context of 
Predictive Maintenance. RCA can be decomposed into four steps:
\begin{enumerate}
	\item \emph{Identify} and \emph{describe} the problem clearly.
	\item Establish a \emph{timeline} from the normal situation
		until the failure finally occurs, through
		the \emph{Point of Incipient Failure}.
	\item Distinguish between the \emph{root cause} and 
		other causal factors. 
	\item Establish a \emph{causal graph} between the root 
		cause and the observed problem.
\end{enumerate}

The trigger signal of the RCA is given by the failure timestamp 
(i.e., the point in time when the failure variable is observed).
Then, RCA yields a list of potential root cause variables
along with their probabilities, which aligns with the way
complex systems fail~\citep{Cook00}. The variables that comprise 
the data are required to be representative enough to help the 
developers and engineers pinpoint the source of the observed problems
through the root causes and their effects~\citep{Weidl08}.

\subsection{Structural Causal Model}
\label{subSCM}

The causal links among the variables $X$ that build the model of
a system are assumed to be most effectively represented using the 
tools from the field of Causality. In this sense, the Structural Causal 
Model (SCM) is the framework that can most generally capture such 
directed associations~\citep{Pearl19}. The SCM defines a set of assignments
governing their specific functional associations $f$, along with 
some independent noise $N$ that accounts for everything that is not 
explicitly included in the model:
\begin{equation}
\label{eqSCM}
	X_j := f_j(PA_j, N_j)\ ,
\end{equation}
where $PA_j$ represents the direct causes of the $X_j$ variable.

If enough knowledge and experience from the field is available
from the subject matter experts,
i.e., strictly complying with the RCA requirements, 
then a complete SCM may be developed right from the start. 
However, this is not the 
typical use-case scenario in complex industrial settings, 
and data generally needs to be carefully leveraged 
to drive the development of the causal model.

\subsubsection{Causal Discovery}
\label{subsubCausDisc}

Whenever the structure of the model
is to be inferred from the observed variables, 
i.e., the Causal Discovery task, assumptions need to be made
about the data generating process, constraints need to be applied, and 
usually the statistical methods of the algorithms yield different 
graphs that explain the same factual data~\citep{Glymour19}. 

In a multivariate environment, the most straightforward approach
is led by the so-called ``constraint-based'' discovery methods.
These traditional approaches iteratively build the causal graph by 
utilizing a score such as the $p$-value of conditional independence 
tests. As a general technique, the Peter-Clark (PC) algorithm is 
described~\citep{Spirtes01}. PC is a causal network structure
learning algorithm that copes well with high dimensionality and can 
often also identify the direction of contemporaneous 
links~\citep{Runge19Nature}. It is
consistent under i.i.d. sampling assuming no latent confounders, i.e.,
all relevant variables need to be observed in the data.
Its outcome is a Markov Equivalence Class, and thus it is likely to have
different graphical representations that explain the same observed 
data. The PC algorithm is especially suited to discover causality in
combination with the Fisher-Z independence test because it requires
less constraints for the input data~\citep{Kobayashi17}.

\subsubsection{Causal Bayesian Network}
\label{subsubCBN}

Once the structural graph that binds the variables is determined, the 
functional associations of the SCM may be learned, and this work adopts a
stochastic interpretation of the world. Therefore, it treats
all $X$ as random variables, and the resulting SCM statistically 
describes their (conditional) probability distributions.

Considering $n$ random variables $X_1,X_2,...,X_n$ and a
directed acyclic graph that relates them causally, a Causal
Bayesian Network (CBN) is a generative model that has the following
factorized joint probability distribution:
\begin{equation}
\label{eqBN}
	P(\mathbf{X}) = P(X_1,...,X_n) = \prod_{j=1}^{n} P \left ( X_j | PA_j, N_j \right ) \ .
\end{equation}

The graphical nature of Bayesian networks allows seeing relationships 
among different variables, and their conditional dependencies
enable performing probabilistic inference~\citep{Alaeddini11}.
Specifically, CBN are powerful tools for knowledge representation 
and inference under \emph{uncertainty}~\citep{Pourret08}.

\subsection{Causal Inference}
\label{subCI}

Beyond probabilistic inference, Causal Inference provides the tools 
that allow estimating causal conclusions even in the absence of a 
true experiment, given that certain assumptions are fulfilled. These 
assumptions increase in strength as is defined in Pearl's Causal 
Hierarchy (PCH) abstraction~\citep{Bareinboim22}, which is
summarized as follows.

\subsubsection{PCH Rung 1: Associational}
\label{subsubPCH1}
Describes the observational distribution of the factual data through
their joint probability function $P(X)$. From this point forward,
interesting quantities, i.e., the queries $X_Q$, can be directly 
computed given some evidence $X_E$, through their conditional
probability:
\begin{equation}
\label{eqQE}
	P(X_Q | X_E) = \frac{P(X_Q,X_E)}{P(X_E)} \ .
\end{equation}

This level of analysis displays a degree sophistication akin to 
classical (un)supervised Machine Learning techniques. As such, it is
subject to confounding bias.

\subsubsection{PCH Rung 2: Interventional}
\label{subsubPCH2}
Describes an actionable distribution, which endows causal information
at the population level.
This level of analysis can be achieved through actual experimentation
via Randomized Control Trials, or through statistical adjustments
that smartly combine observed conditional probabilities to 
reduce spurious associations in the estimation. Pearl's 
$do$-calculus is likely to be the most effective approach to 
determine the identifiability of causal effects by applying the
following three rules: 1) insertion/deletion of observations, 2)
action/observation exchange, and 3) insertion/deletion of 
actions~\citep{Pearl12}.

\subsubsection{PCH Rung 3: Counterfactual}
\label{subsubPCH3}
Describes a potential distribution at the individual level
driven by hypothetical speculations over data that may contradict 
the facts. Conducting this estimation requires the following 
three steps~\citep{PearlCIS}:
\begin{enumerate}
\item {\bf Abduction}: Beliefs about the world are initially
updated by taking into
account all the evidence $E$ given in the context. Formally, the 
exogenous noise probability distributions $P(U)$ are updated to $P(U|E)$.
\item {\bf Action}: Interventions are then conducted to reflect the 
counterfactual assumptions, and a new causal model is thus created.
\item {\bf Prediction}: Finally, counterfactual reasoning occurs over 
the new model using the updated knowledge.
\end{enumerate}

\section{Method}
\label{secMethod}

Since the applied industrial environment belongs to the area of Predictive
Maintenance, the observation of a common development standard such as the
ISO 13374 is recommended~\citep{iso13374}. This specification breaks down the 
complexity of a problem into small modules that may be developed 
in isolation, thus increasing the chances of project success while
improving the interpretability and explainability of the technical
solution, and also help to reduce the technical debt.
This section describes the Data Manipulation, State Detection,
and Health Assessment processing blocks.

\subsection{Data Manipulation}
\label{subDataMan}
Causality is an emergent property of complex industrial 
systems~\citep{Yuan24}.
In this setting, \emph{event} variables constitute high level, nominal,
time-stamped, qualitative data records that group functions into 
categories and hierarchies. In fact, the causal relation is a
relation among events (not properties or states)~\citep{Bunge09}.
To preprocess these collected event 
logs~\citep{VanHoudt22}, a message template extractor is
typically used~\citep{Chuah10}.

\subsubsection{Event Transformation}
\label{subsubEventTransf}
To progress with their analysis, the event variables need to be standardized
into a time-series format through a transformation~\citep{Hu07}. 
A common approach is to utilize a counting function, which
adds up the number of logged messages within a given time-slot,
therefore yielding an integer-valued representation for all the
variables.
Other details such as the sampling rate need to match the speed
of change of the variables. Finally, the data need to be
windowed on the (recent) past of the observed failure, considering a limited
history in time, which needs to be sufficient to observe the
evolution of the health condition of the system (from normal to failure).
Once the failure has occurred, the state of the system is assumed
to change as operators may take actions to mitigate its impact and
prevent further damage~\citep{LiRCA22}.

\subsubsection{Relevance Filter}
\label{subsubRelevFilt}
Since the event-variable space may be large at this stage, 
it is advised to reduce it by filtering the relevant variables only.
In this sense, the proposed data strategy consists of first using 
a robust measure of Mutual Information between any two
variables~\citep{Reshef16}, then discarding the
non-significant variable relationships via independence testing, and
finally ranking the remaining variables. At last, it is also advised to
remove periodic events such as timers that are unrelated to 
troubleshooting, e.g., using Fourier analysis and 
regressions~\citep{Kobayashi17}. As a result, a collection
of significant integer-valued time series variables representing the
evolution of event counts is obtained.

\subsection{State Detection}
\label{secStateDetect}
This module builds the data-driven SCM and conducts a
\emph{coarse-grained}
diagnosis by determining if the asset under test shows a normal or 
abnormal working condition. However,
if the resolution is not adequate in the sampled data, 
the causal precedence may not be observed, leading to cycles 
and unobserved confounding. Therefore, what makes this approach 
especially suited for dynamic data is the explicit 
consideration of time in the causal model, which is especially 
required to break cycles and resolve race conditions.

\subsubsection{Structure Learning}
\label{subsubStructLearn}
The proposed approach initially infers the causal relations from observational 
time series event data. Nevertheless, no family or method for causal 
discovery in time series stands out in all situations with 
different characteristics~\citep{Assaad22}.

An initial baseline is obtained with the PC algorithm
(using the Fisher-Z independence test)
on data augmented with time lags. The count-based transformation 
described in Section~\ref{subsubEventTransf} naturally lends
itself to the application of this technique as long
as the counts approximate a Gaussian distribution,
which can be asserted
using the Lilliefors normality test~\citep{Lilliefors67}.

However, the direct application of PC discovery may not be advised 
for certain time series cases, and other more involved methods 
using more powerful statistical tests (also with time lags) 
should be explored on top of 
it. In consequence, the Momentary Conditional Independence 
(PC-MCI) test is considered~\citep{Runge19}, which
has a stronger causal detection ability based on partial 
correlation tests.

\subsubsection{Dynamic Networks}
\label{subsubDynCBN}
What follows is the construction of the probabilistic model from the 
learned time-dependent causal structure. This approach is agnostic to 
any specific (non/linear) parametric functional relationship, and also 
provides natural access to its inherent uncertainty (even at the 
individual instance level).

In this sense, the Dynamic CBN (DCBN) yield a factorized 
representation of a stochastic process. They extend the standard 
causal Bayesian network formalism by providing explicit
discrete temporal dimensions. DCBN represent a probability distribution 
over the possible histories of a time-invariant process; their advantage 
with respect to classical probabilistic temporal models like a Markov 
chain is that a DCBN is a stochastic transition model factored over 
a number of random variables, over which a set of conditional dependency 
assumptions is defined~\citep{Bobbio08}.

Considering $n$ time-dependent discrete random variables 
$X_1^t,X_2^t,...,X_n^t$, a DCBN is essentially their replication
over time slices $t-\Delta$ (creating the so-called
discretization steps), with the addition of a set of arcs in the graph
representing the transition model, which is defined through the
distribution $P(X_i^{t}|X_j^{t-\Delta})$, for all time-related
variables $i$ and $j$. Arcs connecting nodes at different time-slices 
($\Delta > 0$) are called interslice edges, while arcs connecting 
nodes at the same slice ($\Delta = 0$) are called intraslice edges.
The joint probability distribution of the DCBN is shown as follows:
\begin{equation}
\label{eqDCBN}
	P(\mathbf{X}) = \prod_{\forall j} P \left ( X_j^{t-\Delta} | PA_j^{t-\Delta} \right ) \ .
\end{equation}

\subsubsection{Failure Prediction}
\label{subsubFailPred}
The learned DCBN shall be used to estimate the probability of
the Failure variable $X_F$ in time $P_F(t)$, which is the sink
node in the model that represents the eventual system crash,
given the observed data (i.e., the root causes and their effects):
\begin{equation}
\label{eqPF}
	P_F(t) = P(X_F^{t} | PA_F^{t}) \ .
\end{equation}

Ideally, the probability of observing a high count of failure 
events $X_F^{t}=H$ should be a monotonically increasing function (in
time) until the moment of system failure.

To detect if an anomaly is present, the Point of Incipient
Failure $T$ should be determined. This is
the moment in time when the system \emph{starts} developing an 
abnormal behavior that will eventually lead to the crash. Also,
this is where the root cause of the observed anomaly is 
reasonably expected to be found.
A possible strategy to determine this instant can be defined 
by the minimum-time significant-second-derivative of the probability of 
Failure, as this is the \emph{first} inflection point with a
\emph{minimally relevant} increase $\Theta$ of risk (it may not
be the greatest absolute increase, but it shall be one with
the precedence in time):

\begin{equation}
\label{eqPOIF}
	T = \min_{t} \left ( \frac{\partial^2}{\partial t^2} P_F(t) > \Theta \right ) \ .
\end{equation}

While there may be many different ways to express this criterion,
by using a $\Theta$ threshold parameter the subject matter experts
can be easily involved in the design of the solution.

\subsection{Health Assessment}
\label{subHealthAss}

This module exploits the probabilistic SCM and conducts a
\emph{fine-grained}
diagnosis by determining the root cause of the observed anomaly.

\subsubsection{Path Finding}
\label{subsubPathFind}
The hypothesis of isolation is a methodological requirement of the
sciences for research; hence, the useful fiction of the isolated
``causal chain'' or ``singled-out path'' in the structure
will work to the extent to which such an isolation
takes place, and this is often the case in definite respects during
limited intervals of time. Moreover, since every isolable process is causal,
anomalies can emerge solely as a result of external
perturbations~\citep{Bunge09}.

Concerning the analysis of a DCBN for RCA, estimating the most 
likely time-sequence chain of variables for the observed anomaly 
event adds explanatory value in an industrial environment.
In the DCBN, each node represents an event count or state 
change of a variable, and the arcs represent causal–temporal 
relationships between the nodes. In this setting,
probabilistic temporal logic determines that causes and effects
are steady state formulas, the properties of which hold for the 
system at a certain point in time~\citep{VanHoudt22}, and this 
allows for each formula to be a path formula too where multiple 
variables are involved.
Therefore, the causal paths shall be given by the structure of the 
graph: a search algorithm shall be used to traverse it and find 
all the routes $S$ from the different root nodes to the sink
Failure node.

For the Point of Incipient Failure $T$, the most likely causal 
path $S^{*}$ that explains the anomaly data can be
determined after the exhaustive search among all the potential
paths $S$ and their respective probabilities:

\begin{equation}
\label{eqLikelyPath}
	S^{*} = \max_{s \in S} P(s | \hat s) ;\ t = T \ ,
\end{equation}
where $s$ represents a structural path from a source node
to the sink node (i.e., the failure event variable).

Conditioning on the variables not in the path under analysis
($\hat s$) is important to block spurious associations. This is
especially relevant in the case of descendants, because in the event
of an anomaly, the parent/ancestor variables are preferred as
precedents~\citep{LiRCA22}.

Finally, 
in addition to putting the focus on the most expected behavior, 
one could argue that the root cause may also have occurred
in the most unexpected/irregular setting~\citep{Yang22},
assuming that the most commonly experienced issues will have already been solved.
This alternative perspective may also be covered in the
proposed approach by minimizing the path likelihood probability.

\subsubsection{Algorithmic Recourse}
\label{subsubCFRCA}
So far, the main focus of the analysis has been on the observed 
factual data. However, these data represent 
only one of the many potential outcomes the system could have 
experienced: had things been different, an alternative outcome
may have been observed. Algorithmic Recourse explores these
counterfactual worlds~\citep{Karimi22}. Such environments
are simulated via inference through (atomic) interventions 
$\alpha$ in 
time on a specific abnormal instance in order to revert the 
anomaly~\citep{Han23}, i.e., to lower the risk of failure $X_F$.
This is expected to help in the recognition and understanding 
of the general root causes that lead to the system 
failure~\citep{LiRCA22}.

Formally, the specific retrospective reasoning that these
counterfactuals explore on the anomaly, i.e., the Point
of Incipient Failure at $t=T$, can be stated as:
\begin{equation}
\label{eqAlgoRec}
	P(X_F^{t=T} = L | do(X^{t=T}=\alpha), X^{t=T}, X_F^{t=T}=H) \ .
\end{equation}

Given that an anomaly was factually recorded in the data, i.e.,
through observing a high risk of failure $X_F^{t=T}=H$,
i.e., a high count of failure events,
Equation~(\ref{eqAlgoRec}) estimates the probability that
the risk would have been low at the Point of 
Incipient Failure $X_F^{t=T}=L$, had the root cause $X^{t=T}$
had the value $\alpha$, instead of the value it actually had
when the anomaly was triggered. Note that this formula does
not involve regular probabilistic conditioning, but the
application of the Abduction-Action-Prediction process
described in Section~\ref{subsubPCH3}.

\section{Results}
\label{secResults}

This section elaborates on the experimental work. For further
details, the code along with the description of the required
software tools is available 
here.\footnote{Code repository: https://github.com/atrilla/ci4ts}

\subsection{System Data Description}
\label{subSysDataDesc}
\begin{figure}[!htb]
  \centering
  \begin{tikzpicture}[shorten >=1pt,on grid,auto,node distance=1.5cm] 
   \node[state] (x1)  []  {$X_1$};
   \node[state] (x2)   [right=of x1] {$X_2$};
   \node[state] (x3)   [right=of x2] {$X_3$};
   \node[state] (y)   [below=of x2] {$Y$}; 

   \path[->] 
        (x1) edge (x2)
        (x2) edge (x1)
        (x1) edge [bend left=45] (x3)
        (x3) edge [bend right=45] (x1)
        (x2) edge (x3)
        (x3) edge (x2)
        (x1) edge (y)
        (x2) edge (y)
        (x3) edge (y)
        ;
\end{tikzpicture}
\caption{Time-implicit summary graph for a synthetic system that could describe 
	a 2-out-of-3 redundancy. Note the time-confounded associations
	among the $X$ channels.}
\label{fig2oo3}
\end{figure}
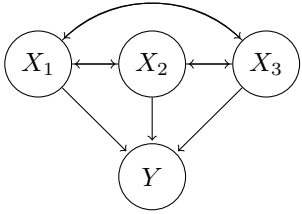

For illustrative purposes, the system considered in this work is
synthetic. It is comprised of 4 integer-valued time-series variables
that could describe a 2-out-of-3 redundant system as follows:
three full-duplex data channels that exchange messages among the devices $X$,
and three simplex alarm channels from $X$ to $Y$, which checks that the system
is in good working condition, see Figure~\ref{fig2oo3}.

\begin{figure}[!htb]
  \centering
  \includegraphics[width=0.9\linewidth]{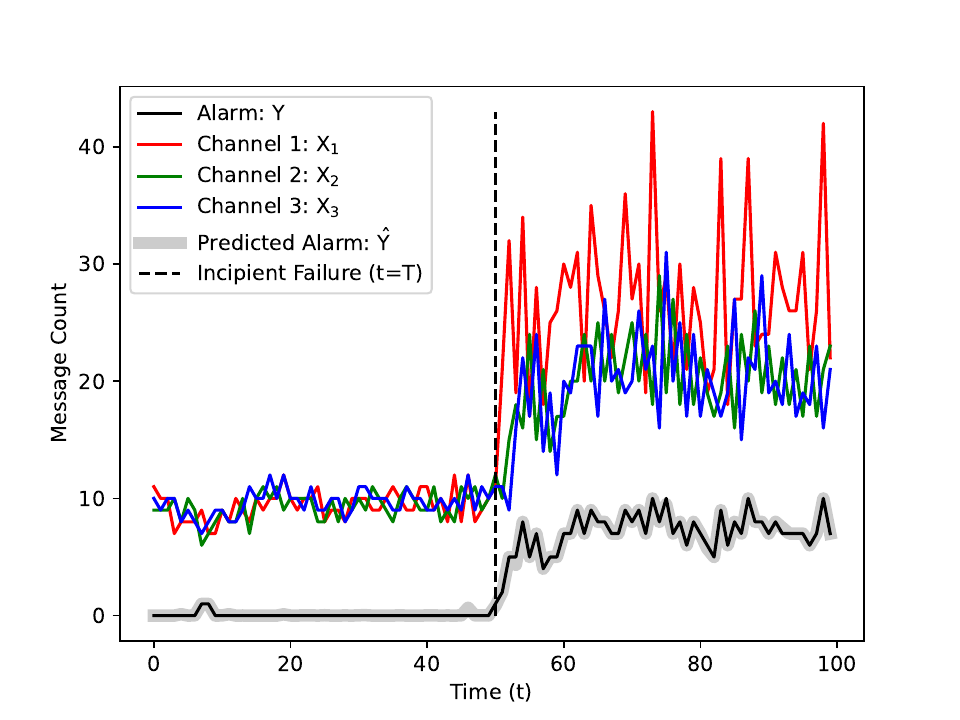}
\caption{Timeline of system condition evolution showing a failure on
	Channel 1.}
\label{figDataPlot}
\end{figure}

The variables represent the counts of event messages over time.
The window of analysis frames the timeline evolution of the 
system condition, from a normal operation to the crash, through 
the Point of Incipient Failure, where the root cause is likely to
be found, see Figure~\ref{figDataPlot} for a specific example
of an anomaly on Channel 1. The failure is simulated as an 
injection of exogenous noise into one device $X$, which then
propagates to the rest of the devices.
This simplified environment should account for all of the
data collection and manipulation steps.

\subsection{Causal Model Learning}
\label{subCausModLearn}

\begin{figure}[!htb]
  \centering
  \begin{tikzpicture}[shorten >=1pt,on grid,auto,node distance=1.5cm] 
  \node[state] (x3t2)  []  {$X_3^{t-2}$};
  \node[state] (x2t1)   [right=of x3t2] {$X_2^{t-1}$};
  \node[state] (x1t2)   [right=of x2t1] {$X_1^{t-2}$};
  \node[state] (x1t1)   [right=of x1t2] {$X_1^{t-1}$};
  \node[state] (x3t1)   [right=of x1t1] {$X_3^{t-1}$};
  \node[state] (x1t)   [below=of x2t1] {$X_1^{t}$};
  \node[state] (x3t)   [below=of x1t2] {$X_3^{t}$};
  \node[state] (x2t)   [below=of x1t1] {$X_2^{t}$};
  \node[state] (yt)   [below=of x3t] {$Y^{t}$}; 

   \path[->] 
        (x3t2) edge (x1t)
        (x2t1) edge (x1t)
        (x2t1) edge (x3t)
        (x1t2) edge (x3t)
        (x1t1) edge (x2t)
        (x3t1) edge (x2t)
        (x1t) edge (yt)
        (x3t) edge (yt)
        (x2t) edge (yt)
        ;
\end{tikzpicture}
\caption{Time-explicit ground truth graph.} 
\label{figStructGT}
\end{figure}
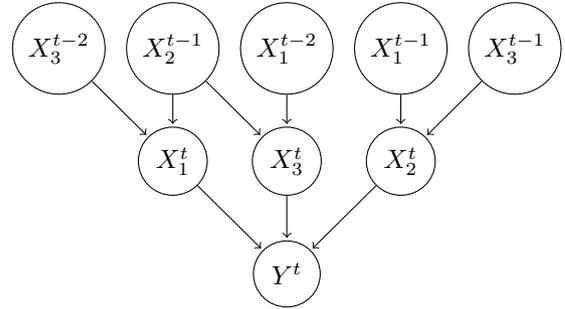

The structure of the model is learned with time-dependent
discovery algorithms to deal with the implicit confounding issue.
Figure~\ref{figStructGT} shows the ground truth of the
time-explicit graph of the system under analysis. 
In this structure, the lagged terms correspond to the
channel data, and the contemporaneous terms correspond
to the alarm signal.

To statistically train the causal model, 
a dataset of 100 instances is generated,
barely over the required minimum amount of failure examples for the
three potential root cause variables~\cite{Lejeune10}.
Table~\ref{tabCausDiscPerf} charts the causal discovery
performance of PC (stable version,
manually augmented with time lags) and PC-MCI
algorithms over different $p$-value configurations, using the 
Structural Hamming Distance, which compares the
resulting graphs by computing the difference between their 
(binary) adjacency matrices.
According to these results, the simplicity of the PC algorithm
augmented with manual time lags like a buffer, which is a
necessary step to fairly compare the results (PC-MCI automatically
implements this), seems to be more effective to discover the
structure that binds the variables of the synthetic system
(all results seemed reasonable, e.g., there were no oddities
such as empty graphs).

\begin{table}
    \centering
    \caption{Performance of causal discovery algorithms through the
	Structural Hamming Distance (SHD).}
    \label{tabCausDiscPerf}
    \begin{tabular}{rl}
      \toprule 
	    \bfseries Algorithm ($p$-value) & \bfseries SHD\\
      \midrule 
	    PC-MCI (0.01) & 20\\
	    PC-MCI (0.03) & 20\\
	    PC-MCI (0.05) & 21\\
      \midrule 
	    PC-stable (0.01) & 9\\
	    PC-stable (0.03) & 9\\
	    PC-stable (0.05) & 10\\
      \bottomrule 
    \end{tabular}
\end{table}

This performance analysis on learning the structural information
has been conducted for research informative purposes only.
The ground truth graph is used from this point
forward to train the probability distributions of the DCBN model.
The performance of this
predictive model is evaluated using the root mean square error
metric compared with the alarm variable, and it yields a score 
of $0.1247$. In the illustrative timeline example shown in
Figure~\ref{figDataPlot}, the predicted alarm is
almost indistinguishable from the actual alarm signal.

\subsection{Causal Diagnosis}
\label{subCausModLearn}

Having a model that is able to accurately predict the observed
alarm level, and therefore, the anomaly at the Point 
of Incipient Failure $T$ before the issue propagates,
Table~\ref{tabPaths} shows the list of paths that traverse its
graph from a source (channel) node to the sink (alarm) node, 
ranked by their likelihood scores.
Note how the most likely path explains the observed anomaly at $t=T+$,
immediately after the incipient failure has occurred on Channel 1
(this diagnosis is necessarily reactive since by definition
it cannot be anticipated).

\begin{table}
    \centering
    \caption{Likelihood-ranked paths that explain the observed 
	anomaly right after the Point of Incipient Failure $t=T+$.}
    \label{tabPaths}
    \begin{tabular}{rl}
      \toprule 
	    \bfseries Causal Path & \bfseries Likelihood\\
      \midrule 
	    $X_2^{t-1} \rightarrow X_1^{t} \rightarrow Y^{t}$ & $0.0505$\\
	    $X_2^{t-1} \rightarrow X_3^{t} \rightarrow Y^{t}$ & $0.0293$\\
	    $X_3^{t-2} \rightarrow X_1^{t} \rightarrow Y^{t}$ & $0.0208$\\
	    $X_1^{t-2} \rightarrow X_3^{t} \rightarrow Y^{t}$ & $0.0124$\\
	    $X_3^{t-1} \rightarrow X_2^{t} \rightarrow Y^{t}$ & $0.0080$\\
	    $X_1^{t-1} \rightarrow X_2^{t} \rightarrow Y^{t}$ & $0.0076$\\
      \bottomrule 
    \end{tabular}
\end{table}

\begin{figure}[!htb]
  \centering
  \includegraphics[width=0.9\linewidth]{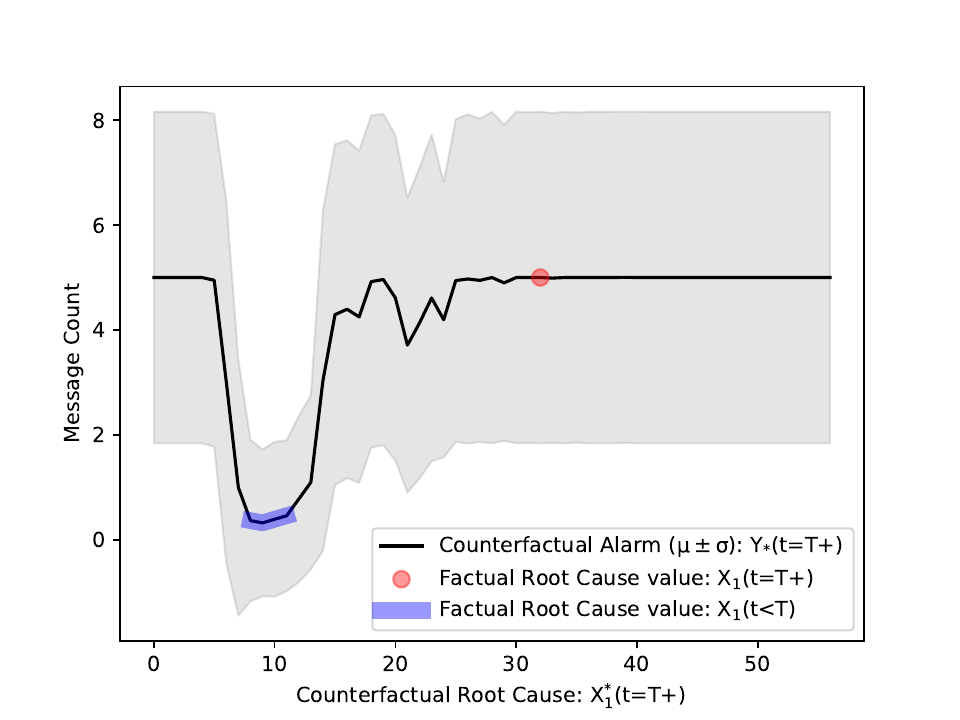}
	\caption{Distributions of potential alarm outcomes for 
	a range of counterfactuals on the root cause channel.
	The descriptive averages in terms of means and standard
	deviations for the (discrete) failure random variable are shown.}
\label{figCF}
\end{figure}

Finally, to further assert the blame for the root cause, the 
following counterfactual is evaluated 
$P(Y_{*}^{t=T+} | do(X_1^{t=T+}), X_1^{t=T+}, Y^{t=T+})$.
The resulting estimand shall adjust for the anticausal
backdoor path introduced by the $X_2^{t-1}$ confounder.
Figure~\ref{figCF} shows the results for a range of potential
alarm outcomes through their distributions.
Note how the alarm level would have stayed low had the root
cause $X_1^{t}$ kept the values it had before the Point of Incipient 
Failure. Also note how the region around those values is the one
that displays the least amount of uncertainty. In retrospect,
the diagram also shows how this one single variable, i.e. the
root cause, was sufficient to cause the anomaly that would
end up with the system failure.

\section{Discussion}
\label{secDisc}
Up to this point, the solution presented in this workshop paper
has described the basic principles of its causal RCA technology 
and an initial experimental proof 
of concept has been shown. This early stage of maturity corresponds
to a standard ISO 16290 Technology Readiness Level (TRL) 
of 3~\citep{ISOTRL}.
This section brainstorms some avenues of improvement to increase
this robustness indicator up to higher quality standards,
considering the specific challenges of complex industrial environments,
especially towards pinpointing the incipient failure,
which is where the root cause is most likely to be found.

\subsection{Validate the technology in a relevant environment (TRL 4--5)}
\label{subTRL45}
One first idea could be to improve the learning of the structure 
of the model. In line with similar constraint-based approaches, the
consideration of tiers that heuristically stratify groups of
variables can be advantageous~\citep{Andrews20}. Alternatively,
score-based approaches, where multiple candidate models are fit
and checked, can also be explored~\citep{Glymour19}. Finally, the
usage of reductionist Functional Causal Models (FCM) 
may be especially helpful. A FCM represents a pairwise (or bivariate) 
interaction of the effect as an \emph{analytic function} of a direct 
cause and some unmeasurable noise. Several forms of the FCM have 
been shown to be able to produce unique causal directions and have 
received practical applications~\citep{Huang15}.
In specific scenarios such as multivariate time series, FCM
can even improve the performance of traditional 
constraint-based approaches~\citep{Runge19Nature}.
Additionally, different independence tests can be introduced in the 
FCM-based discovery process to tackle heterogeneous data settings.

Second, in industrial settings where \emph{only} the crash event at the
end of the timeline is available (instead of an evolving alarm
signal), a different approach shall be 
adopted. If a similar solution based on regression is still pursued, 
a transform based on artificially prepending the failure with a 
``rise-time'' pattern driven by prior experience could be 
introduced~\citep{Hu07}.
Alternatively, a classification approach based on logistic
regression could be adopted by treating the single crash event as 
a binary target variable, but data imbalance may be a concern.
Finally, the kink discontinuity at the Point of Incipient Failure
or the sharp discontinuity at the Failure could
be exploited with a regression-based difference-in-differences 
technique~\citep{Abadie05}, provided that data are available 
after the critical event, which may not always be the case.

Finally, assessing the capacity to scale of the described solution
is necessary. This goal should consider the impact of the number of 
variables, the length of the records, the nature of data, etc.

\subsection{Demonstrate the technology in an operational environment (TRL 6--7)}
\label{subTRL67}
The next enhancement idea has to do with the
processing of real-world data, and while they may come from the 
lab (perhaps also using an accelerated degradation testing 
procedure), what is actually required are data from the field. 
However, operational data often suffer from imbalance issues, 
especially showing a
shortage of failure instances. In this case, probably the most sensible way
forward is to adopt causal approaches that initially tackle the 
detection of anomalies, in line with the standard pipeline of
Predictive Maintenance, including point, contextual, and collective 
irregularities. 
Causality-based anomaly detection methods provide at least 
two significant theoretical benefits over purely statistical 
methods: 1) improved robustness to non-anomalous out-of-distribution 
data, which implies a reduction in false-alarms, and 2) a 
potential for failure localization due to the topological 
ordering of the causal graph~\citep{Strelnikoff23}.

What is potentially weak in the current state
of the art in causal RCA is the specific assessment of 
heteroskedasticity. The causal relationships are stationary 
unless an anomaly occurs~\citep{Yang22}.
Anomalous data are non-stationary, and this 
violates one of the fundamental assumptions of time series models. 
Therefore, a stronger emphasis on preprocessing transformations may
be necessary. Additionally, the case of time-varying exposure 
in the presence of time-varying confounders requires 
special attention~\citep{Hernan23}. 

Finally, it would also be interesting to relax the assumption that
failures can only occur through the path of the root cause, and
explore the impact of direct and indirect effects. In this case,
mediation analysis is a specific application of counterfactuals 
that seeks to identify and explain the mechanism or 
process that underlies an observed relationship between an independent 
variable (i.e., the root cause) and a dependent variable (i.e., the failure
effect) via the inclusion of a third variable, known as 
a mediator variable, an intermediary variable, or an intervening 
variable~\citep{VanderWeele16,Agler17}.

\section{Conclusion}
\label{secConc}
This workshop paper has developed a complete top-down
counterfactual Root Cause Analysis approach from first causal
inference principles that is also compliant with industrial 
development guidelines. On the basis of processing
multivariate time series data, the focus of this diagnosis challenge 
has been put on detecting 
the Point of Incipient Failure, which displays the first anomaly 
pattern before the issue propagates to the rest of the system.
This moment in time is believed to be where the root cause is 
most likely to be found. This hypothesis has been illustrated
on a synthetic system showing how one single counterfactual
is sufficient to explain the anomalous behavior, therefore
pinpointing the root cause of the eventual failure problem.

\bibliography{ci4ts2024-dialogs}

\end{document}